\pdfoutput=1

\documentclass[11pt]{article}

\usepackage{emnlp2021}

\usepackage{times}
\usepackage{latexsym}

\usepackage[T1]{fontenc}

\usepackage[utf8]{inputenc}

\usepackage{microtype}
\usepackage{amsthm}
\usepackage{amsmath}
\usepackage{amssymb}
\usepackage{bm}
\usepackage{bbding}
\usepackage{multirow}
\usepackage{makecell}
\usepackage{booktabs}
\usepackage{threeparttable}
\usepackage{subfig}
\usepackage[pdftex]{graphicx}
\graphicspath{{figures/}}
\title{Fine-grained Semantic Alignment Network for Weakly Supervised Temporal Language Grounding}

\author{Yuechen Wang \quad Wengang Zhou \quad Houqiang Li\thanks{\ \ Corresponding author.} \\
        University of Science and Technology of China \\
        \texttt{wyc9725@mail.ustc.edu.cn, \{zhwg, lihq\}@ustc.edu.cn} \\
    }

\begin{document}
\maketitle
\begin{abstract}
Temporal language grounding (TLG) aims to localize a video segment in an untrimmed video based on a natural language description. 
To alleviate the expensive cost of manual annotations for temporal boundary labels,
we are dedicated to the weakly supervised setting, where only video-level descriptions are provided for training.
Most of the existing weakly supervised methods generate a candidate segment set and learn cross-modal alignment through a MIL-based framework.
However, the temporal structure of the video as well as the complicated semantics in the sentence are lost during the learning.
In this work, we propose a novel \emph{candidate-free} framework: Fine-grained Semantic Alignment Network (FSAN), for weakly supervised TLG.
Instead of view the sentence and candidate moments as a whole, FSAN learns token-by-clip cross-modal semantic alignment by an iterative cross-modal interaction module, generates a fine-grained cross-modal semantic alignment map, and performs grounding directly on top of the map.
Extensive experiments are conducted on two widely-used benchmarks: ActivityNet-Captions, and DiDeMo, where our FSAN achieves state-of-the-art performance. 
\end{abstract}

\section{Introduction}
Given an untrimmed video and a natural language sentence, Temporal Language Grounding~(TLG) aims to localize the temporal boundaries of the video segment described by a referred sentence.
TLG is a challenging problem with great importance in various multimedia applications, \emph{e.g.}, video retrieval~\cite{Shao_2018_ECCV}, visual question answering~\cite{MovieQA,VQA,8988148}, and visual reasoning~\cite{COG}.
Since its first proposal~\cite{TALL,DiDeMo}, tremendous success has been made on this problem~\cite{ijcai2018-143,chen-etal-2018-temporally,Ge2019MACMA,aaai2019Yuan,MAN,He2019ReadWA,cvprWangHW19,2D-TAN,I2N}.
Despite the achievements with supervised learning, the temporal boundaries for every sentence query need to be manually annotated for training, which is expensive, time-consuming, and potentially noisy. 
On the other hand, it is much easier to collect a large amount of video-level descriptions without detailed temporal annotations, since video-level descriptions naturally appear with videos simultaneously on the Internet~(\emph{e.g.,} YouTube).
To this end, some prior works are dedicated to  weakly supervised setting, where only video-level descriptions are provided, without temporal labels.

\begin{figure}[tb]
	\centering
	\includegraphics[scale=0.23]{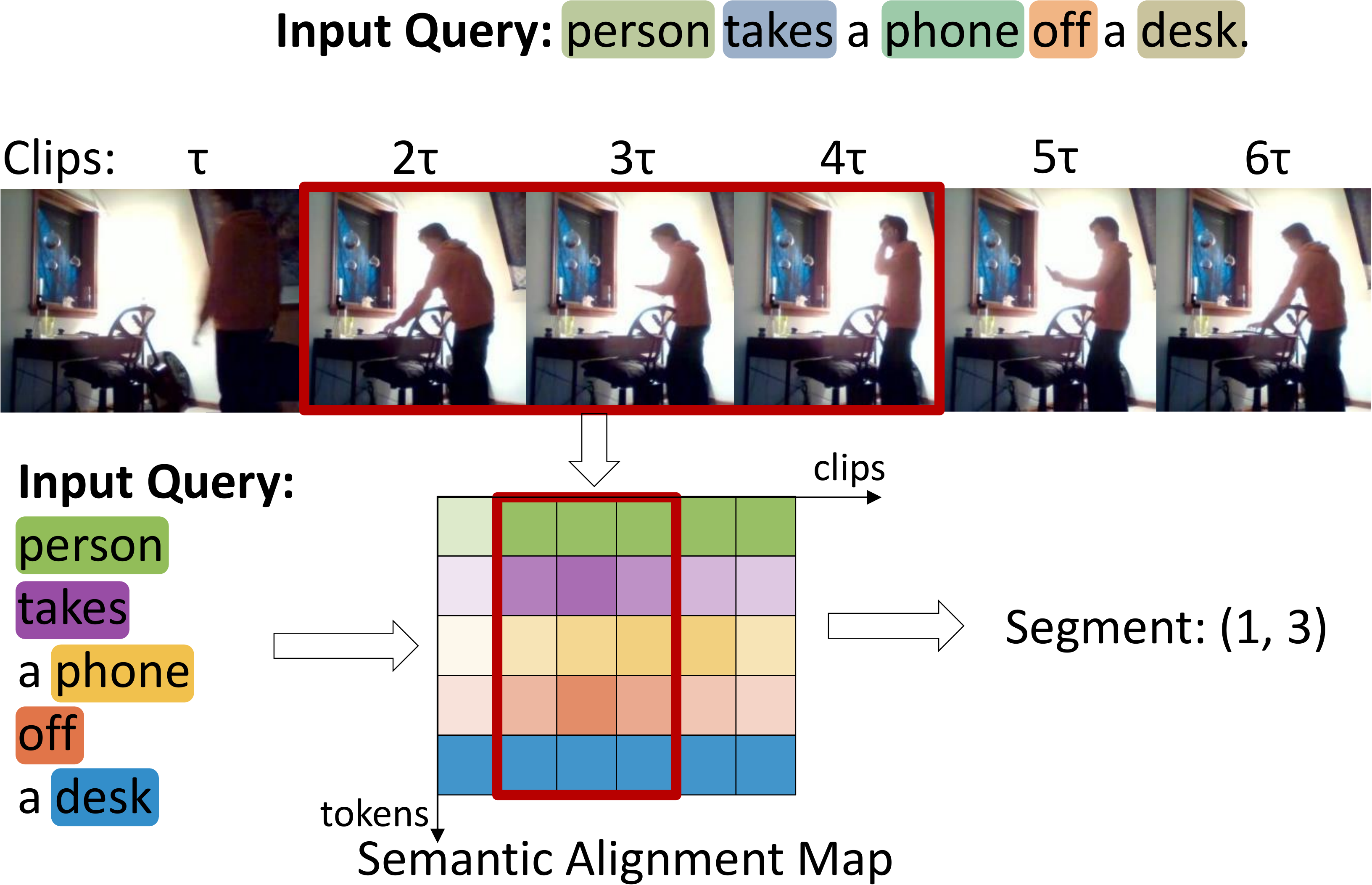}
	\caption{Illustration of fine-grained semantic alignment map for temporal language grounding.
	}
	\label{fig:fig1}
\end{figure}

Most of the previous weakly supervised methods follow a Multiple Instance Learning~(MIL) paradigm, which samples matched and non-matched video-sentence pairs, and learn a matching classifier to implicitly learn the cross-modal alignment. However, during the matching classification, the input sentence is often treated as a single feature query, neglecting the complicated linguistic semantics.
VLANet~\cite{VLANet} treats tokens in the input sentence separately, and performs cross-modal attention on token-moment pairs, where the moment candidates are carefully selected by a surrogate proposal selection module to reduce computation cost. 
SCN~\cite{SCN} proposes to generate and select moment candidates and performs semantic completion for the sentence to rank selected candidates.
Nevertheless, the generation and selection process of moment candidates also involves high computational costs. In addition, the moment candidates are considered separately, while the temporal structure of the video is also important for grounding.
Figure~\ref{fig:fig1} shows an example to localize the query ``\emph{person takes a phone off a desk}'' in the given video. If the model views the sentence as a whole and performs matching classification, it is hard to learn undistinguished words like ``\emph{off}'' during the training.
However, the neglected words may play important roles to determine the temporal boundaries of the described moment.

In this paper, we propose a novel framework named a Fine-grained Semantic Alignment Network (FSAN), for weakly supervised temporal language grounding.
The core idea of FSAN is to learn token-by-clip cross-modal semantic alignment presenting as a token-clip map, and ground the sentence on video directly based on it.
Specifically, given an untrimmed video and a description sentence, we first extract their features by visual encoder and textual encoder independently. 
Then, an Iterative Cross-modal Interaction Module is devised to learn the correspondence between visual and linguistic representations. 
To make temporal predictions for grounding, we further devise a semantic alignment-based grounding module. 
Based on the learned cross-modal interacted features, a token-by-clip semantic alignment map is generated, where the $(i, j)$-th element on the map indicates relevance between the $i$-th token in the sentence and $j$-clip in the video.
Finally, an alignment-based grounding module predicts the grounding result corresponding to the input sentence.

Instead of aggregating sentence semantics into one representation and generating video moment candidates, FSAN learns a fine-grained cross-modal alignment map that helps to retain both the temporal structure among video clips and the complicated semantics in the sentence.
Furthermore, the grounding module in FSAN makes predictions mainly based on the cross-modal alignment map, which alleviates the computation cost of candidate moment representation generation.
We demonstrate the effectiveness of the proposed method on two widely-used benchmarks: ActivityNet-Captions~\cite{ActivityNet} and DiDeMo~\cite{DiDeMo}, where state-of-the-art performance is achieved by FSAN.

\section{Related Work}
\subsection{Temporal Language Grounding}
Temporal language grounding is proposed~\cite{TALL,DiDeMo} as a new challenging task, which requires deep interactions between two visual and linguistic modalities.
Previous methods have explored this task in a fully supervised setting~\cite{TALL, DiDeMo,  chen-etal-2018-temporally, Ge2019MACMA, Xu2019MultilevelLA, aaai19Chen, aaai2019Yuan, SIGIRZhang, MAN, lu-etal-2019-debug}. Most of them follow a two-stage paradigm: generating candidate moments with sliding windows and subsequently matching the language query.
Reinforcement learning has also been leveraged for temporal language grounding~\cite{He2019ReadWA,cvprWangHW19,Cao_2020_MM}.

Despite the boom of fully supervised methods, it is very time-consuming and labor-intensive to annotate temporal boundaries for a  large number of videos. And due to the annotation inconsistency among annotators, temporal labels are often ambiguous for models to learn. 
To alleviate the cost of fine-grained annotation, weakly supervised setting is explored lately~\cite{TGA, WSLLN, SCN, VLANet, ZhangMM}.
TGA~\cite{TGA} exploits maps video candidate features and query features into a latent space to learn cross-modal similarity. 
In~\cite{VLANet}, a video-language attention network is proposed to learn cross-modal alignment between language tokens and video segment candidates.
Differently, our FSAN gets rid of the trouble of generating candidates and learns fine-grained token-by-clip semantic alignment.

\begin{figure*}[htb]
	\centering
	\includegraphics[scale=0.57]{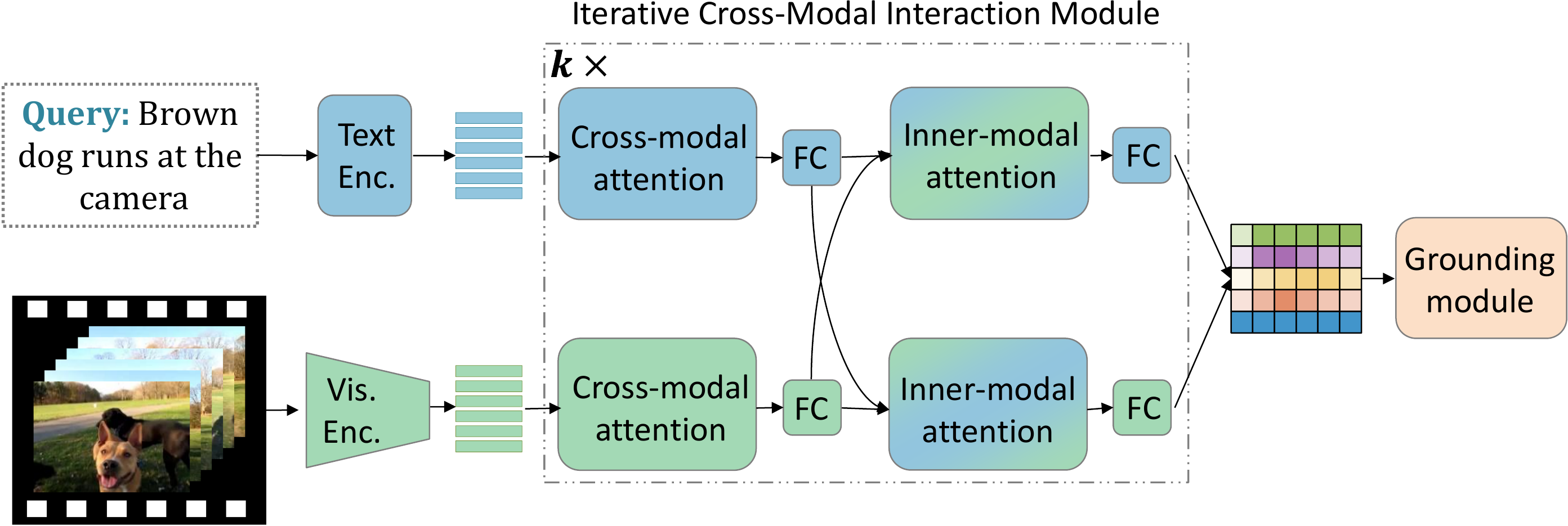}
	\caption{The architecture of FSAN. 
		It consists of four main components: (1) a text encoder, (2) a visual encoder, (3) an iterative cross-modal interaction module, and (4) a proposal module.
	}
	\label{fig:overview}
\end{figure*}

\subsection{Transformer in Language and Vision}
Since it is first proposed by Vaswani \emph{et al}.~\cite{NIPS2017_3f5ee243} for machine translation, transformer has become a prevailing architecture in NLP.
The basic block of transformer is the multi-head attention module, which aggregates information from the whole input in both transformer encoder and decoder module. 
Transformer demonstrates superior performance in language model pretraining methods~\cite{Devlin2019BERTPO, Radford2018ImprovingLU, Yang2019XLNetGA}, and achieves competitive performance on diverse NLP problems.
Recently, transformer has been introduced to various computer vision tasks, such as image classification~\cite{Chen2020GenerativePF}, image generation~\cite{Parmar2018ImageT}, object detection~\cite{DETR}, semantic segmentation~\cite{wang2021maxdeeplab}, tracking~\cite{Wang2021TransformerMT}, \emph{etc.} Comparing to CNN, the attention mechanism learns more global dependencies, therefore, transformer also shows great performance in low-level tasks~\cite{chen2020pretrained}. Transformer has also been proved effective in multi-modal area, including multi-modal representations~\cite{DeVLBert, LXMERT, VLBERT, VideoBERT} and applications~\cite{BotianCaptioning, XinchengTransformer, LiangSemi}. 
Inspired by the great success, we devise an iterative cross-modal interaction module mainly based on the multi-head attention mechanism.

\section{Our Approach}

Given an untrimmed video and a text-sentence query, a temporal grounding model aims to localize the most relevant moment in the video, represented by its beginning and ending timestamps. 
In this paper, we consider the weakly supervised setting, \emph{i.e.}, for each video $V$, a textual query $S$ is provided for training. The query sentence describes a specific moment in the video, yet the temporal boundaries are not provided for training.
In the inference stage, the weakly trained model is required to predict the beginning and ending timestamps of the video moment that corresponds to the input sentence $S$.

We present a novel framework named Fine-grained Semantic Alignment Network (FSAN) for the temporal language grounding problem. 
As shown in Figure~\ref{fig:overview}, given a video and text query, we first encode them separately. 
The resulting representations then interact with each other through an iterative cross-modal interaction module. 
The outputs are used to learn a Semantic Alignment Map (SAP) between the two modalities.
Finally, the SAP is fed into an alignment-based grounding module to predict scores for all possible moments.

In the following subsections, we will first introduce the visual and language encoder, then describe the Iterative Cross-Modal Interaction Module. Finally, we will elaborate on the semantic alignment map and the grounding module based on it.


\subsection{Input Representation}
\textbf{Language Encoder. }
We use a standard transformer encoder~\cite{NIPS2017_3f5ee243} to extract the semantic information for the input query sentence $S$. Each token in the input query is first embedded using GloVe~\cite{Glove}. The resulting vectors are mapped to dimension of $d_s$ by a linear layer and fed into a transformer encoder to obtain context-aware token features $\bm{S} = \{\bm{w}_i\}_{i=1}^{N_s}$, where $N_s$ is the number of tokens and $\bm{w}_k \in \mathbb{R}^{d^s}$ denotes the feature of $k$-th token in the sentence.

\noindent\textbf{Video Encoder. }
For the input videos, we extract visual features using a pretrained feature extractor and then apply a temporal pooling on frame features to divide it into $N_v$ clips. Hence the video can be represented by $\bm{V} = \{\bm{v}_j\}_{j=1}^{N_v}$, where $\bm{v}_j \in \mathbb{R}^{d^v}$ denotes the feature of $j$-th video clip, and $d_v=d_s$ is the dimension of visual feature. 
Experimental results illustrate that the computation cost is considerably reduced by the temporal pooling.

\subsection{Iterative Cross-Modal Interaction Module}
Inspired by the great success of transformer encoder on vision-language pretraining~\cite{UnicoderVL,LXMERT}, we devise an Iterative Cross-modal Interaction Module~(ICIM) to learn the semantic relevance between visual and textual representations.
The module is composed of a stack of 6 layers, and each layer consists of cross-modal attention, inner-modal attention, and feed-forward layers.
The core component for cross-modal interaction is the multi-head attention, which is also vital in the transformer structure. Formally, given two sequences of $d$-dimensional features $X=[\bm{x}_1, \cdots, \bm{x}_{N_x}]$ and $Y=[\bm{y}_1, \cdots, \bm{y}_{N_y}]$, the calculation of multi-head attention is as follows:
\begin{align}
	\begin{split}
		\bm{Q}_i =\bm{W}_i^Q \bm{X},& \bm{K}_i=\bm{W}_i^K \bm{Y}, \bm{V}_i=\bm{W}_i^V \bm{Y}, \\
		\bm{A}_i =& softmax(\frac{(\bm{Q}_i \bm{K}_i^T)}{\sqrt{d}})\bm{V_i},
	\end{split} \\
	MA(\bm{X}, \bm{Y}) &= \bm{W}^M(\bm{A}_1||\bm{A}_2||\cdots ||\bm{A}_m),
	\label{equ:attention}
\end{align}
where $A_i$ is output the $i$-th of $m$ attention heads. The final output $MA(X, Y)$ of multi-head attention is of same dimension as the input $X$.

For the textual representation $\bm{S}\in\mathbb{R}^{N_x\times d^S}$ and visual representation $\bm{V}\in\mathbb{R}^{N_x\times d^V}$ input to the iterative cross-modal interaction module, we first adopt cross-modal attention, \emph{i.e.},
\begin{align}
	\begin{split}
	\bm{S}^{'} &= LN(MA(\bm{S}, \bm{V})+\bm{S}), \\
	\bm{S}^{''} &= LN(FFN(\bm{S}^{'})+\bm{S}^{'}),
	\label{equ:crossattn_1}
	\end{split} \\
	\begin{split}
	\bm{V}^{'} &= LN(MA(\bm{V}, \bm{S})+\bm{V}), \\
	\bm{V}^{''} &= LN(FFN(\bm{V}^{'})+\bm{V}^{'}),
	\label{equ:crossattn_2}
	\end{split}
\end{align}
where $LN$ denotes layer normalization and $FFN$ denotes feed-forward layer. 
To retain the temporal structure of the video and the grammar of the sentence, we add learnable positional encodings to the input of each modality.
Through the above attention operation, the features of different modalities are able to freely interact with the other modality to learn a fine-grained semantic alignment.

To model the inner-modal context after cross-modal interaction, we further apply an inner-modal attention, which is similar with the calculation in Equation~\eqref{equ:crossattn_1} and \eqref{equ:crossattn_2}, except that the multi-head attention is applied only on single-modal representation, \emph{i.e.}, self-attention on single-modal features.
After 6 iterations of cross-modal interaction and inner-modal modeling, the enhanced features $\bm{S}^{''}$ and $\bm{V}^{''}$ are fed into subsequent proposal module to predict a cross-modal semantic alignment map, and perform grounding based on it.

\subsection{Semantic Alignment Map \protect\footnote{Several mistakes are corrected from the conference version.}}
After iterative cross-modal and inner-modal attention in ICIM, the correspondence can be fully explored between each pair of textual tokens and video clips. Therefore, a token-by-clip Semantic Alignment Map (SAP) $\bm{P}$ with size $N_s \times N_v$ can be learned for temporal grounding. 
Formally,
\begin{equation}
\begin{aligned}
	\bm{P} &= SAP(V, S) \\
			&= Norm(\bm{W}^s{\bm{S}^{''}}^T)\cdot Norm(\bm{W}^v{\bm{V}^{''}}^T),
\end{aligned}
\end{equation}
where $\bm{W}^s \in \mathbb{R}^{d_l\times d_s}$ and $\bm{W}^v \in \mathbb{R}^{d_l\times d_v}$ are learnable parameters, $\cdot$ denotes dot product, and $Norm(\cdot)$ denotes L2-normalization. 

The value of the $(i, j)$-th element on the SAP $\bm{P}_{ij}$ represents the relevance between the $i$-th textual token and the $j$-th video clip.
However, the SAP can learn the cross-modal relationship only with supervision that indicates semantic alignment. To this end, we adopt a video-level matching loss, which is calculated as:
\begin{equation}
\begin{aligned}
	\mathcal{L}^{t} = &\max(0,\delta-S(V, S)+S(V, S^{-}),
	\label{equ:losstri}
\end{aligned}
\end{equation}
where $S^{-}$ is a non-matching description sentence randomly sampled from the dataset, and $S(V, S)$ is the matching score function between video $V$ and sentence $S$, defined as:
\begin{equation}
	S(V, S) = \max_{s\in[0,N_v], e\in[0,N_v]}SC_{s,e},
\end{equation}
which indicates the response between the most relevant moment and the input sentence, and the $SC_{s,e}$ is computed by the following grounding module (Section ~\ref{grounding_module}).

Some same textual expressions may appear in both the positive description and the negative one, confusing the model in the matching classification procedure.
To this end, we mask the repeated tokens on the semantic alignment map $\bm{P}^{-}$ of the input video $V$ and the sampled sentence $S^{-}$.

\subsection{Alignment-Based Grounding Module}
\label{grounding_module}
The elements on the semantic alignment map indicate relevance between video clips and textual tokens, which leads to the idea of a fine-grained alignment-based grounding module. 
The core idea is that if a specific clip $\bm{v}_i$ is part of the described moment $\bm{V}_{s,e}$, where $\bm{V}_{s,e}$ denotes the video segment from the $s$-clip to $e$-th clip. The semantic of $\bm{v}_i$ tends to be highly relevant with all tokens in the description. While if $\bm{v}_i$ is out of the correct moment, at least one token in query is irrelevant to it. Therefore, we score all possible temporal segments formed with video clips by the relevance scores of clips both in and out the segment. 
Considering the clips in the segment as positive clips, and those not in it as negative clips. Then the relevance score of $\bm{V}_{s,e}$ and the query is then defined as:
\begin{equation}
\begin{aligned}
	SC_{s,e} &= Score(\bm{V}_{s, e}) \\
	&= \frac{\sum_{i\in[0,N_s]}\sum_{j\in[s, e]}\bm{P}_{i, j}}{N_s(e-s+1)} \\ &-\frac{\sum_{j\in\complement_[s,e]}\min_{i\in[0, N_s]}\bm{P}_{i, j}}{N_v-(e-s+1)},
	\end{aligned}
\end{equation}
where $\complement_[s,e]$ denotes the aggregation of negative clips, \emph{i.e.}, the complementary set of $\bm{V}_{s,e}$.
The higher the average scores among positive clips for each token, the more possible that $\bm{V}_{[s, e]}$ is relevant to the query. While the lower the average response of negative clips, the less possible that $\bm{V}_{[s, e]}$ is redundant. Through the two-fold filtering, the moment that is more relevant with all tokens in the query will be given a higher score, and therefore more likely to be proposed as the grounding result.

Although the contrastive loss in Equation~\eqref{equ:losstri} enables the model to learn the cross-modal semantic alignment, the temporal discrimination can not be learned under coarse video-level supervision, which is vital for grounding.
To provide fine-level temporal supervision for the fine-grained cross-modal alignment, we further devise a novel two-fold loss on the semantic alignment map $\bm{P}$, including an inner-sample loss and an outer-sample loss.
Specifically, the inner-sample loss aims to enhance the grounded moment on the fine-level alignment map $\bm{P}$. We promote the response among clips in all possible segments, with a weight representing the confidence of the prediction:
\begin{align}
	\mathcal{L}^{i}_{s,e} &= -\frac{\sum_{i\in[0,N_s]}\sigma(\sum_{j\in[s, e]}\bm{P}_{i, j})}{N_s}, \\
	\mathcal{L}^{i} &= \sum_{s\in[0,N_v]}\sum_{e\in[0,N_v]} SC_{s,e}\mathcal{L}^{inner}_{s,e}.
\end{align}
On the other hand, the outer-sample loss aims to suppress complementary part of the video by lowering its weakest response of each clip:
\begin{align}
	\mathcal{L}^{o}_{s,e} &= -\frac{\sum_{j\in\complement_[s,e]}\log(\min_{i\in[0, N_s]}\bm{P}_{i, j})}{N_v-(e-s+1)}, \\
	\mathcal{L}^{o} &= \sum_{s\in[0,N_v]}\sum_{e\in[0,N_v]} SC_{s,e}\mathcal{L}^{outer}_{s,e}.
\end{align}

At the beginning of training, the model is uncertain about the grounding results, therefore, the weight $SC_{s,e}$ varies a little among moment options. As the training continues, the model can give positive moment higher scores easily, hence the loss weight will be larger for positive moments and smaller for negative ones. Therefore, the model will not deviate much from the correct solution.

\subsection{Training and Inference}
The overall training objective is an aggregation of aforementioned losses, given by:
\begin{equation}
\mathcal{L} = \lambda_1\mathcal{L}^{t} + \lambda_2\mathcal{L}^{i} + \lambda_3\mathcal{L}^{o},
\end{equation}
where $\lambda_*$ are hyper-parameters, and satisfy the condition $\lambda_1 + \lambda_2 + \lambda_3 = 1$.

During the inference, the moment with the highest score $SC_{s,e}$ is selected as the grounding result.

\section{Experiments}
\subsection{Datasets and Metrics}
\begin{table}[tb]
	\centering
	\begin{threeparttable}
		\setlength{\tabcolsep}{4pt}{
			\begin{tabular}{l|ccc}
				\toprule[2pt]
				Benchmark & \makecell[c]{Num. of \\ Videos} & \makecell[c]{Num. of \\ Descriptions} & \makecell[c]{Vocab \\ Size} \\
				\hline
				\makecell[l]{ActivityNet-\\Captions} & $14\,950$ & $51\,567$ & $15\,406$ \\
				DiDeMo & $10\,464$ & $41\,206$ & $7\,523$ \\
				Charades-STA & $6\,670$ & $16\,128$ & $1\,289$ \\
				\bottomrule[2pt]
			\end{tabular}
		}
		\caption{Statistics of TLG benchmarks.}
		\label{tab:comp_data}
	\end{threeparttable} 
\end{table}

\begin{table}[tb]
	\centering
	\begin{threeparttable}
		\begin{tabular}{l|ccc}
			\toprule[2pt]
			Method & R@1 &R@5 & mIoU \\
			\hline
			\hline
			\multicolumn{4}{c}{fully supervised methods} \\
			\hline
			MCN~\citeyearpar{DiDeMo} & 28.10 & 78.21 & 41.08 \\
			TGN~\citeyearpar{chen-etal-2018-temporally} & 28.23 & 79.26 & 42.97 \\
			MAN~\citeyearpar{MAN} & 27.02 & 81.70 & 41.16 \\
			I$^2$N~\citeyearpar{I2N} & 29.00 & 73.09 & 44.32 \\
			\hline
			\multicolumn{4}{c}{weakly supervised methods} \\
			\hline
			Random & 3.75 & 22.5 & 22.64 \\
			TGA~\citeyearpar{TGA} & 12.19 & 39.74 & 24.92 \\
			WSLLN~\citeyearpar{WSLLN} & \textbf{19.40} & 53.10 & \underline{25.40} \\
			VLANet~\citeyearpar{VLANet} & 19.32 & \textbf{65.68} & 25.33 \\
			\hline
			\textbf{FSAN} & \textbf{19.40} & \underline{57.85} & \textbf{31.92} \\
			\bottomrule[2pt]
		\end{tabular}
		\caption{Performance comparison on DiDeMo dataset.
			The best and second best numbers are highlighted in \textbf{bold} and \underline{underlined}, respectively.}
		\label{tab:res_did}
	\end{threeparttable}
\end{table}

\textbf{ActivityNet-Captions.}
The ActivityNet-Captions dataset~\cite{ActivityNet} is developed based on ActivityNet dataset~\cite{7298698} ,which contains $20\,000$ untrimmed videos and corresponding language descriptions.
The released ActivityNet-Captions dataset contains $37\,421$ moment-description pairs for training, and $17\,505$, $17\,031$ in val\_1 and val\_2 sets, respectively. Following SCN~\cite{SCN}, we use val\_1 as validation set and val\_2 as test set.

\noindent\textbf{DiDeMo.} 
The Distinct Describable Moments~(DiDeMo) dataset is first proposed in~\cite{DiDeMo}.
It contains over 10k videos selected from Flickr, and all of them are trimmed to a maximum of 30 seconds and equally divided into six 5-second segments. 
Therefore, there are only 21 possible moment candidates for each video.
The DiDeMo dataset is randomly split into training, validation and test set containing $33\,005$, $4\,180$ and $4\,021$ video-sentence pairs, respectively.
For each video, at least 4 annotators are assigned to label text description boundaries.

We exclude the Charades-STA~\cite{TALL} dataset because of its limited scale. 
Charades-STA~\cite{TALL} contains about 6k videos, and the contents are mainly indoor activities.
However, as shown in Table~\ref{tab:comp_data}, comparing to other datasets, Charades-STA is limited in terms of total video amount, number of video-sentence pairs, and vocabulary size. The vocabulary size is critical to enriching the linguistic semantics, hence the semantic diversity is limited in the dataset.

\noindent\textbf{Evaluation Metrics.}
We follow the settings of previous methods~\cite{TGA, SCN}.
For the ActivityNet-Captions dataset, we report results for intersection-over-union~(IoU)$\in$\{0.5,0.3,0.1\} and Recall@\{1,5\}. 
On the DiDeMo dataset, considering the limited number of candidates~(21) and variance among different annotators, we measure the performance with metrics: Rank@1, Rank@5, and mean intersection over union (mIoU). Here Rank@$k$ means the percentage of samples where ground truth moment labeled by different annotators are on average ranked higher than $k$. Following~\cite{DiDeMo}, we discard the worst-ranked ground truth label to reduce the influence of outliers.

\begin{table*}[tb]
	\centering
	\begin{threeparttable}
		\begin{tabular}{l|ccc|ccc}
			\toprule[2pt]
			\multirow{2}*{Method} & \multicolumn{3}{c|}{R@1} & \multicolumn{3}{c}{R@5} \\
			& IoU=0.1 & IoU=0.3 & IoU=0.5 & IoU=0.1 & IoU=0.3 & IoU=0.5 \\
			\hline
			\hline
			\multicolumn{7}{c}{fully supervised methods} \\
			\hline
			ABLR~\citeyearpar{Yuan2019ToFW} & 73.30 & 55.67 & 36.79 & - & - & - \\
			DEBUG~\citeyearpar{lu-etal-2019-debug} & - & 55.91 & 39.72 & - & - & - \\
			CMIN~\citeyearpar{SIGIRZhang} & - & 63.61 & 43.40 & - & 80.54 & 67.95 \\
			2D-TAN~\citeyearpar{2D-TAN} & - & 58.75 & 44.05 & - & 85.65 & 76.65 \\
			\hline
			\multicolumn{7}{c}{weakly supervised methods} \\
			\hline
			Random & 38.23 &  18.64 &  7.63 & 75.74 & 52.78 & 24.49 \\
			WSLLN~\citeyearpar{WSLLN} & \underline{75.4} & 42.8 & 22.7 & - & - & - \\
			SCN~\citeyearpar{SCN} & 71.48 & \underline{47.23} & \underline{29.22}& \underline{90.88} & \underline{71.45} & \underline{55.69} \\
			\hline
			\textbf{FSAN} & \textbf{78.45} & \textbf{55.11} & \textbf{29.43} & \textbf{92.59} & \textbf{76.79} & \textbf{63.32} \\
			\bottomrule[2pt]
		\end{tabular}
		\caption{Performance comparison on ActivityNet-Captions dataset. 
			The best and second best numbers are highlighted in \textbf{bold} and \underline{underlined}, respectively.
			And ``-'' means the result on the metric is not reported in the original paper.
		}
		\label{tab:res_act}
	\end{threeparttable} 
\end{table*}

\begin{table*}[tb]
	\centering
	\begin{threeparttable}
		\begin{tabular}{l|ccc|ccc|c}
			\toprule[2pt]
			\multirow{2}*{Settings} & \multicolumn{3}{c|}{R@1} & \multicolumn{3}{c}{R@5} & \multirow{2}*{mIoU} \\
			& IoU=0.1 & IoU=0.3 & IoU=0.5 & IoU=0.1 & IoU=0.3 & IoU=0.5 & \\
			\hline
			\hline
			w/o grounding module & \textbf{82.58} & 47.99 & 21.09 & 86.34 & 52.60 & 22.85 & 34.57\\
			\hline
			w/o loss on SAP & 65.59 & 46.08 & 26.31 & 88.92 & 66.11 & 36.72 & 30.15 \\
			\hline
			w/o cross-modal attention & 74.51 & 43.36 & 27.03 & 81.02 & 65.37 & 53.41 & 32.68 \\
			w/o inner-modal attention & 76.75 & 50.40 & 28.71 & 91.39 & 71.89 & 50.81 & 34.18 \\
			\hline
			\textbf{full model}  & 78.45 & \textbf{55.11} & \textbf{29.43} & \textbf{92.59} & \textbf{76.79} & \textbf{63.32} & \textbf{36.10} \\
			\bottomrule[2pt]
		\end{tabular}
		\caption{Ablation studies on ActivityNet-Captions dataset.}
		\label{tab:abltaion}
	\end{threeparttable} 
\end{table*}

\subsection{Implementation Details}
For fair comparison, we utilize released visual features as previous methods~\cite{TGA, SCN}. For videos in ActivitiNet-Captions, we adopt C3D~\cite{C3D} features. For DiDeMo, we adopt VGG~\cite{VGG} features. Note that we report the performance of baseline models using the same features as ours. 
The dimension of these features are reduced from 4096 to 500 using PCA. The loss weights in to train FSAN are set to 1/3 equally. The hidden dimensions are set to 512 for all datasets.
We adopt adam algorithm with an initial learning rate of 0.0001. The batch size is set to 128 for all datasets, and the dropout rate is set to 0.1. The code of FSAN is implemented in pytorch, and is trained on one RTX 3090 GPU.

\subsection{Comparisons with State-of-the-art Methods}
We compare the proposed FSAN with multiple baselines, including 1) recently published fully supervised state-of-the-arts methods: MCN~\cite{DiDeMo}, ABLR~\cite{Yuan2019ToFW}, DEBUG~\cite{lu-etal-2019-debug}, CMIN~\cite{SIGIRZhang}, 2D-TAN~\cite{2D-TAN}, TGN~\cite{chen-etal-2018-temporally}, MAN~\cite{MAN}, I$^2$N~\cite{I2N}; and 2) some representative weakly supervised methods: TGA~\cite{TGA}, SCN~\cite{SCN}, WSLLN~\cite{WSLLN}, VLANet~\cite{VLANet}.

\noindent\textbf{Experiments on DiDeMo.}
Table~\ref{tab:res_did} illustrates the performance comparisons on the DiDeMo dataset. 
It can be observed from the numbers that FSAN outperforms TGA and WSLLN on all three metrics. And comparing to VLANet, FSAN performs overall better, except for R@5. This may be due to the surrogate proposal selection module introduced in VLANet~\cite{VLANet}, which in fact performs a two-stage candidate selection and gets rid of temporally overlapped candidates. 

\noindent\textbf{Experiments on ActivityNet-Captions.}
Table~\ref{tab:res_act} illustrates the performance comparisons on the ActivitiNet-Captions dataset. It can be observed that FSAN surpasses previous weakly supervised methods on all metrics. Especially, the FSAN gains about 7\% relative improvement on R@1 with IoU=0.1 over SCN. This is because the average length of videos in the ActivityNet-Captions dataset is relatively long, and the candidate generation-selection-criterion framework of SCN suffers from the limited number of candidates.

\begin{figure*}[tb]
	\subfloat[]{
		\begin{minipage}[b]{0.88\linewidth}
			\includegraphics[scale=0.64]{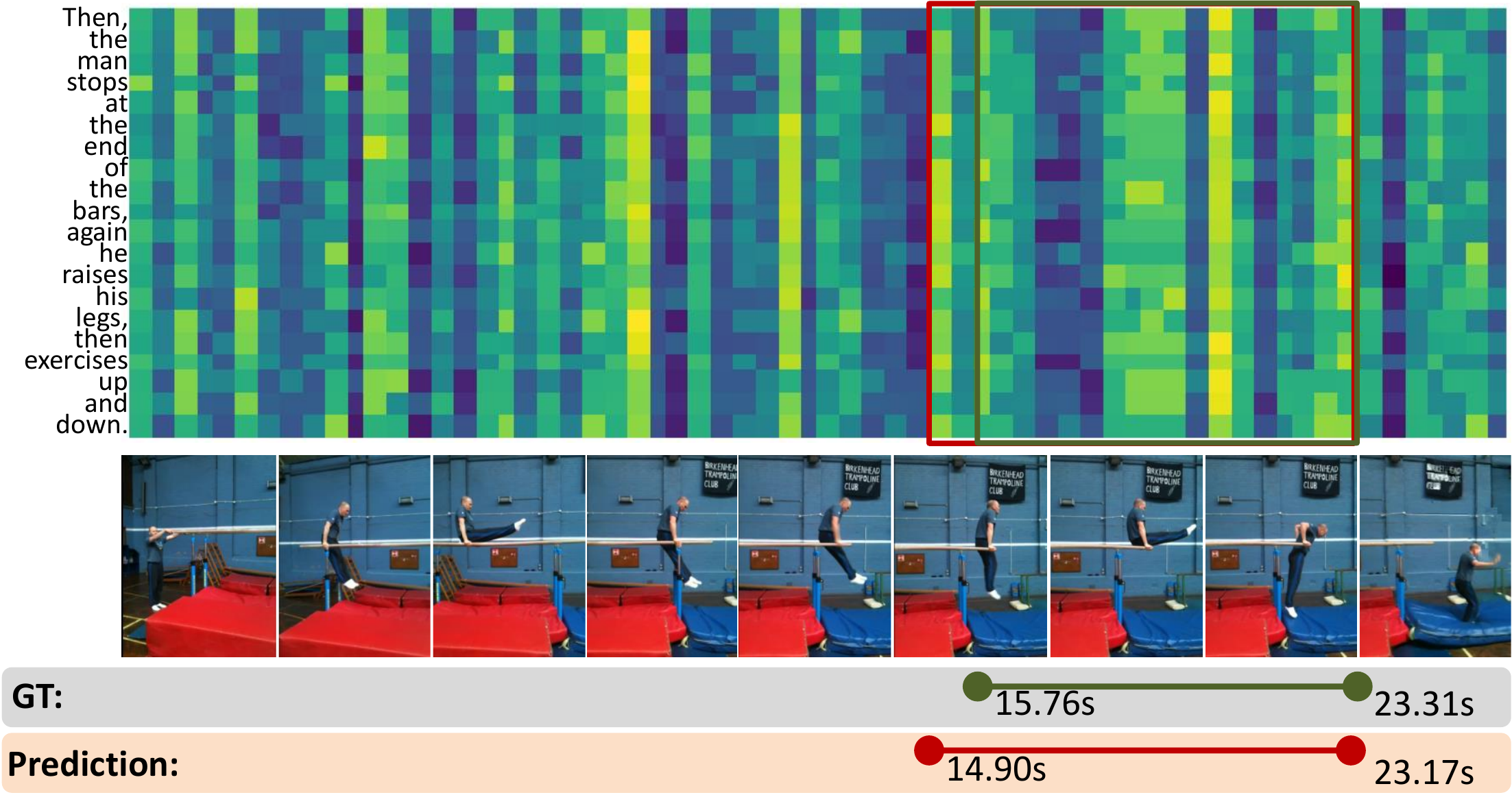}
		\end{minipage} 
		\hfill
		\begin{minipage}[b]{0.1\linewidth}
			\centering	
			\includegraphics[scale=0.63]{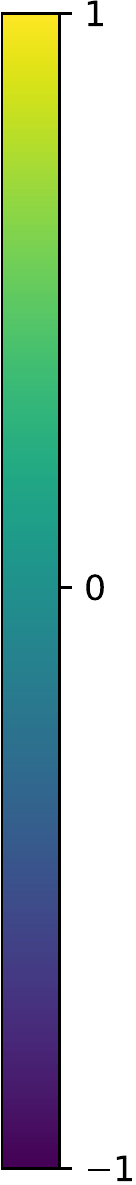}
		\end{minipage}
	}
	
	\subfloat[]{
		\begin{minipage}[b]{0.88\linewidth}
			\includegraphics[scale=0.64]{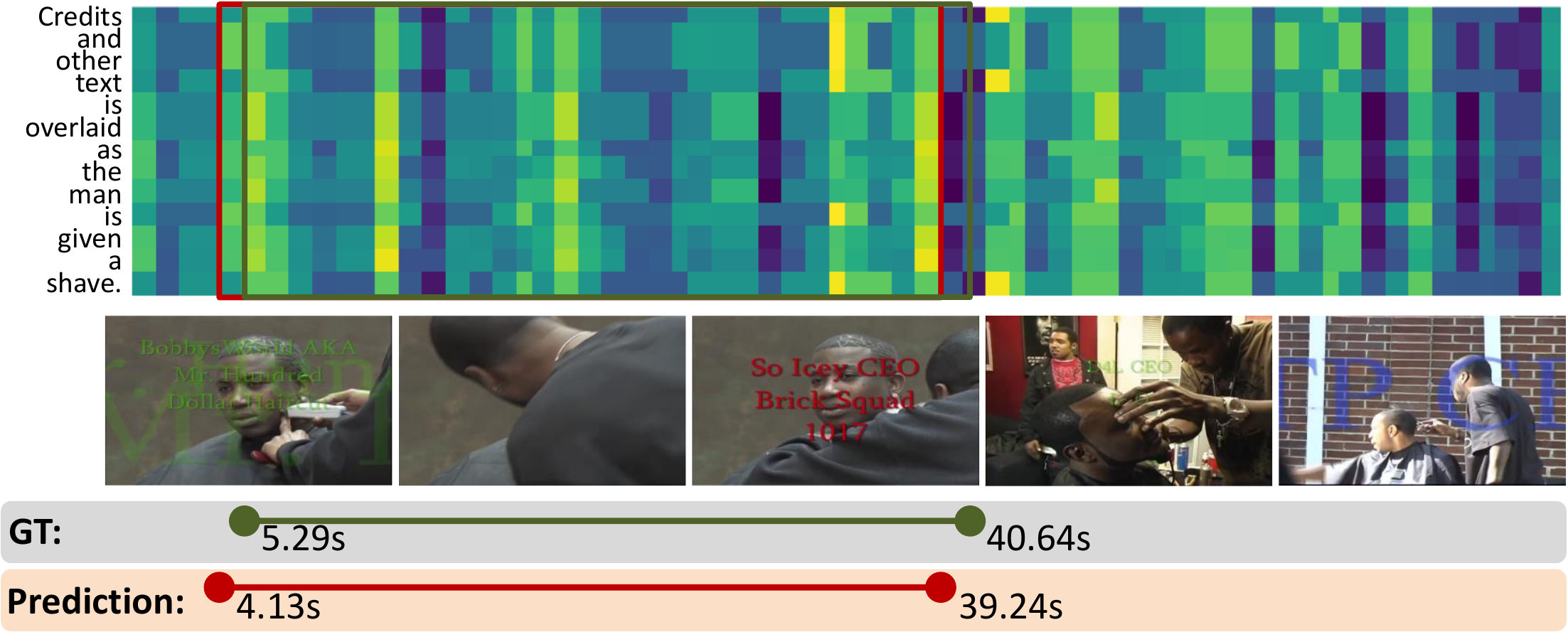}
		\end{minipage} 
		\hfill
		\begin{minipage}[b]{0.1\linewidth}
			\centering	
			\includegraphics[scale=0.63]{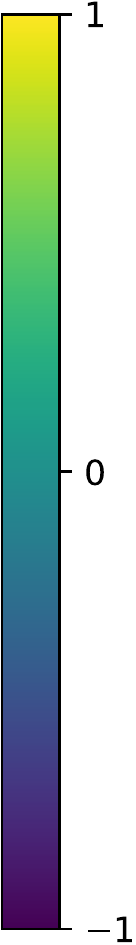}
		\end{minipage}
	}
	\caption{Visualization of the semantic alignment map and grounding results by FSAN.
		The rows and columns on the map correspond to text tokens and video clips, respectively. 
		The green boxes on the map refer to the ground truth temporal region, while the red boxes refer to output of FSAN.
	}
	\label{fig:result}
\end{figure*}

\begin{figure}[tb]
	\includegraphics[scale=0.18]{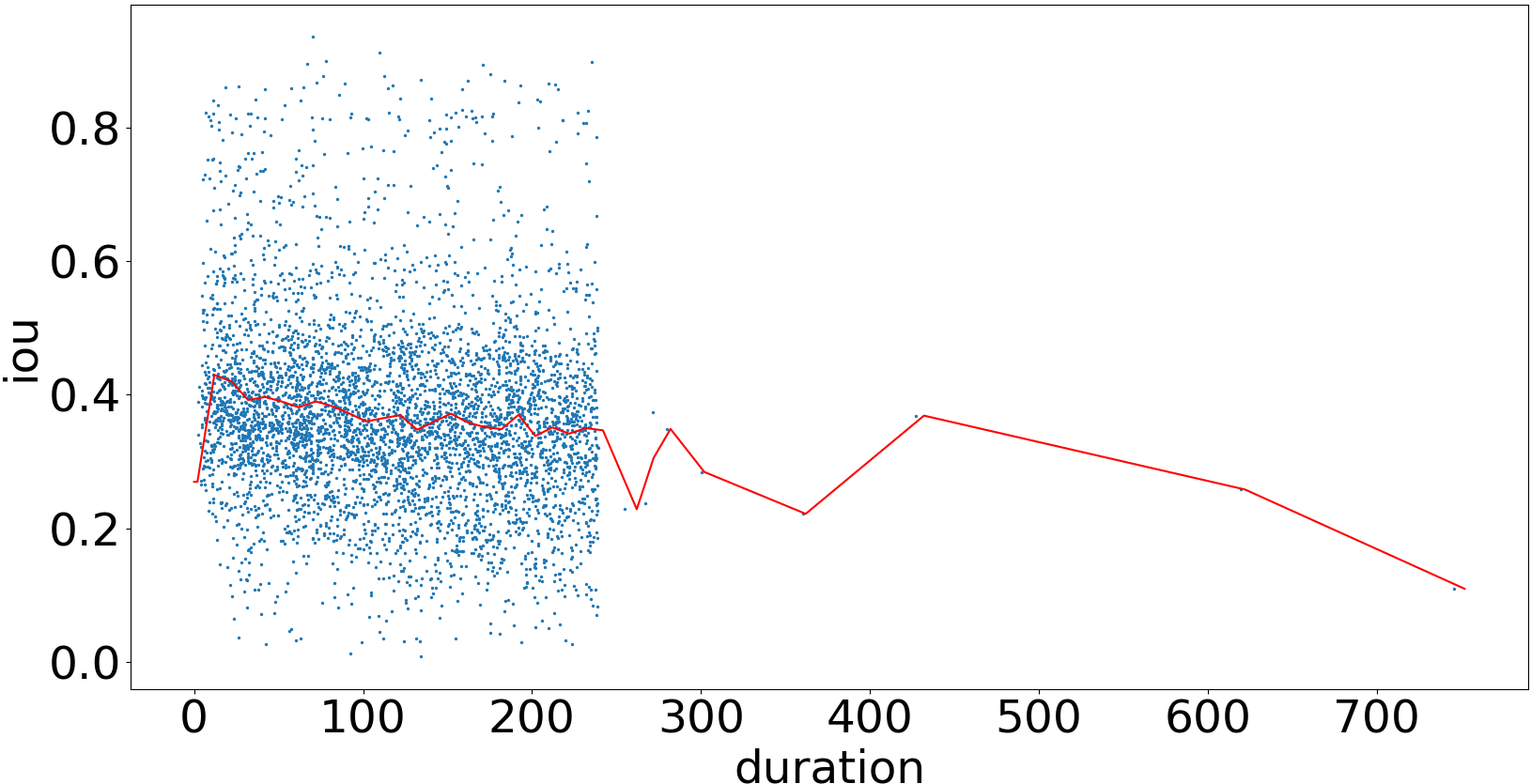}
	\caption{Visualization of the performance variation under different video length.
		blue points indicate individual videos, and the red line indicates the mIoU of videos in the corresponding length interval.}
	\label{fig:analysis}
\end{figure}

\subsection{Ablation Study}
To investigate the importance of each component in FSAN, we conduct ablation experiments. Results are shown in Table~\ref{tab:abltaion} and we give detailed discussions in the next subsections. 
Note that for comparison, mean intersection-over-union~(mIoU) is not reported in the previous subsection, which calculates the average IoU of rank 1$^{st}$ predictions. However, we report and compare mIoU among FSAN variants in this section.

\noindent\textbf{Impact of Grounding Module. }
To validate the effectiveness of the alignment-based grounding module, we devise a common yet competitive prediction layer upon the visual branch output of ICIM. Specifically, we apply an attention pooling on the text-aware visual features $\bm{V}^{''}$, then apply a three-layer MLP to predict matching score for all possible temporal segments.
Results are shown in the 1$^{st}$ row in Table~\ref{tab:abltaion}. It can be observed that without the grounding module based on the semantic alignment map, the performance drops rapidly on strict IoU metrics~(IoU=0.5, 0.3), which demonstrates the temporal precision improvement by introducing the token-by-clip alignment map.

\noindent\textbf{Impact of Loss on SAP. }
The 2$^{nd}$ row in Table~\ref{tab:abltaion} shows the result without inner-sample loss and outer-sample loss. Under this setting, the grounding performance drops on all metrics compared to full FSAN. 
The explanation is that without the two losses refining SAP, the FSAN is trained only by the video-level matching loss $\mathcal{L}^{tri}$.
Hence the model can learn video-level coarse semantic alignment, while neglecting token-wise sentence semantics as well as the temporal structure of the video. 

\noindent\textbf{Impact of ICIM. }
We study the role of cross-modal attention and inner-modal attention in the iterative cross-modal interaction module.
It can be observed in the 3$^{th}$ and 4$^{th}$ rows in Table~\ref{tab:abltaion} that both of them contribute to the grounding performance.
Concretely, without inner-modal attention, mIoU drops by 1.92\%.
While without cross-modal attention, mIoU drops by 3.42\%.
These results demonstrate the importance of both two attention mechanisms in capturing temporal context among video clips, sentence structure among tokens, and token-by-clip cross-modal semantic alignment.

\subsection{Analysis and Visualization}
We also visualize some examples of the grounding result of FSAN in ActivityNet-Captions dataset. as shown in Figure~\ref{fig:result}. For each video-sentence pair, we visualize the token-clip semantic alignment map, as well as the ground truth and predicted temporal boundaries.

In the first example, the description sentence is long and complicated, consisting of three sequential activities~(\emph{stop, raise} and \emph{exercise up and down}). FSAN achieves high IoU~($0.88$) on this difficult case, which indicates the strong ability of FSAN to learn fine-grained semantics from both visual and linguistic modalities.
The second example shows the ability of FSAN to not only detect objects and their actions in video, but also understand abstract descriptions for video (\emph{credits, text}). To better understand abstract descriptions is one of the key points for TLG to develop from action localization. 
In addition, in the second example, the main action \emph{shave} is blocked in some frames, which is challenging for grounding. Though the blocking reflects in the visualized alignment map, FSAN manages to locate the complete moment.

To further analyze the performance of FSAN, we plot a graph showing how the performance varies as video length grows. As shown in Figure.~\ref{fig:analysis}, the performance of FSAN is relatively stable when video length grows, with a trend of weakening.
For example, the mIoU for shortest videos (2-12s, 130 cases) and longest videos (>230s, 140 cases) are 42.99 and 34.68, respectively. 

\section{Conclusion}
In this paper, we present a novel framework for temporal language grounding, namely Fine-grained Semantic Alignment Network~(FSAN). 
To capture fine-level video-language semantic alignment, we devise an iterative cross-modal interaction module, which enables single-modal representations to interact with each other.
Furthermore, we propose to perform temporal grounding based on a semantic alignment map, which alleviates the generation of video candidates.
We conduct experiments on two widely-used benchmarks: ActivityNet-Captions and DiDeMo, and achieve state-of-the-art performance on both datasets, which demonstrates the effectiveness of our proposed FSAN.

\section*{Acknowledgements}
This work was supported by the National Natural Science Foundation of China under Contract 61632019.

\bibliography{anthology,custom}
\bibliographystyle{acl_natbib}

\appendix

%

\end{document}